%% file: icml.tex
\documentclass{article}

\usepackage{microtype}
\usepackage{graphicx}
\usepackage{subfigure}
\usepackage{booktabs} 

\usepackage{hyperref}

 \usepackage[preprint]{icml2026}

\usepackage{amsmath}
\usepackage{amssymb}
\usepackage{mathtools}
\usepackage{amsthm}

\usepackage[capitalize,noabbrev]{cleveref}

\usepackage{adjustbox}
\usepackage{makecell}

\theoremstyle{plain}

\theoremstyle{definition}

\theoremstyle{remark}

\usepackage[textsize=tiny]{todonotes}

\usepackage{multirow}

\icmltitlerunning{OptPO: Optimal Rollout Allocation for Test-time Policy Optimization}

\begin{document}

\twocolumn[
\icmltitle{OptPO: Optimal Rollout Allocation for Test-time Policy Optimization}

\icmlsetsymbol{equal}{*}

\begin{icmlauthorlist}
\icmlauthor{Youkang Wang}{polyu}
\icmlauthor{Jian Wang}{polyu}
\icmlauthor{Rubing Chen}{polyu}
\icmlauthor{Tianyi Zeng}{polyu}
\icmlauthor{Xiao-Yong Wei}{scu,polyu}
\icmlauthor{Qing Li}{polyu}
\end{icmlauthorlist}

\icmlaffiliation{polyu}{Department of Computing, The Hong Kong Polytechnic University}
\icmlaffiliation{scu}{Department of Computer Science, Sichuan University}

\icmlcorrespondingauthor{Xiao-Yong Wei}{cs007.wei@polyu.edu.hk}

\icmlkeywords{Machine Learning, ICML}

\vskip 0.3in
]

\printAffiliationsAndNotice{}  

\begin{abstract}

Test-time policy optimization enables large language models (LLMs) to adapt to distribution shifts by leveraging feedback from self-generated rollouts. However, existing methods rely on fixed-budget majority voting to estimate rewards, incurring substantial computational redundancy. 
We propose Optimal Rollout Allocation for Test-time Policy Optimization (OptPO), a principled framework that adaptively allocates inference budgets. By formulating the voting process as a Bayesian sequential probability ratio test, OptPO dynamically halts sampling once the posterior confidence in a consensus answer exceeds a specified threshold. 
Crucially, it utilizes the retained rollouts for on-policy updates, seamlessly integrating with algorithms like PPO or GRPO without requiring ground-truth labels.
Across diverse reasoning benchmarks, OptPO significantly reduces rollout overhead compared to fixed-sample baselines while preserving or improving accuracy. 
By unifying statistically optimal stopping with test-time learning, OptPO offers a computationally efficient paradigm for test-time adaptation.
The source code will be open upon acceptance at \url{https://open-upon-acceptance}.
\end{abstract}

\input{sections/1-introduction}

\input{sections/2-related-work}
\input{sections/4-methodology}
\input{sections/5-experiments}
\input{sections/6-conclusion}


\bibliography{icml}
\bibliographystyle{icml2026}


\end{document}

%% file: sections/1-introduction.tex
\section{Introduction}
\label{sec:introduction}

Large language models (LLMs) have achieved remarkable advances in natural language understanding, reasoning, and problem solving \citep{brown2020language, achiam2023gpt}. Yet, a persistent challenge emerges when models face distribution shifts at inference time, such as domain-specific terminology, unfamiliar reasoning styles, or linguistic variability, which can cause substantial performance degradation \citep{hu2025test}. To mitigate this, test-time adaptation that addresses these shifts dynamically has thus emerged as a critical research direction.

As a promising paradigm, \emph{test-time learning} (TTL) enables models to learn to adapt during inference. Early methods primarily focused on light-weight adaptation without updating parameters, such as entropy minimization or batch normalization \citep{wang2020tent, niu2022efficient}. More recent work extends these ideas to \emph{test-time training} (TTT), with explicit parameter updates occurring at inference, which substantially enhances task-specific adaptability. For example, \citet{akyurek2024surprising} suggests that test-time training improves few-shot learning under distribution shift, while \citet{hubotter2024efficiently} proposes active fine-tuning that selectively adapts LLMs using informative feedback signals. 


Within the TTT paradigm, a particularly emerging direction is test-time policy optimization. By treating the inference process as a reinforcement learning (RL) problem, methods such as RL with verifiable rewards (RLVR) \citep{wang2025reinforcement} and TTRL \citep{zuo2025ttrl} utilize consensus-based feedback (e.g., majority voting) to guide policy updates. 
These approaches leverage the ``wisdom of the crowd'' from multiple rollouts to construct label-free rewards, effectively bridging the gap between reasoning consistency and correctness.
However, a critical bottleneck remains: \textit{rollout inefficiency}. Standard test-time RL methods typically mandate a fixed, large number of rollouts for every query to estimate rewards robustly. 
This ``one-size-fits-all'' strategy is computationally wasteful, as easy instances establish consensus quickly, while only hard instances require extensive sampling. 
Consequently, the heavy inference cost limits the practical deployment of test-time policy optimization.


This paper introduces \emph{Optimal Rollout Allocation for Test-time Policy Optimization} (\textbf{OptPO}), a general framework designed to address the efficiency-accuracy trade-off. 
Our key insight is to reframe reward estimation as a sequential hypothesis testing problem. Instead of pre-committing to a fixed sample budget, OptPO adaptively estimates the vote distribution using a Bayesian Sequential Probability Ratio Test (SPRT). 
Sampling halts as soon as the posterior evidence for a consensus answer satisfies a user-defined error tolerance. 
The accumulated rollouts are then immediately repurposed for policy optimization (e.g., via PPO \citep{schulman2017proximal} or GRPO \citep{shao2024deepseekmath}), ensuring that compute is allocated only where necessary to reduce uncertainty.
Furthermore, our OptPO can be broadly compatible with supervised fine-tuning for test-time training.

In summary, our contributions are as follows:
\begin{itemize}
    \item We identify the fixed-budget rollout mechanism as a primary source of inefficiency in current test-time policy optimization approaches.
    \item We propose OptPO, a principled, plug-and-play framework for test-time policy optimization that adaptively allocates rollouts based on sequential majority voting with early stopping.
    \item We demonstrate that OptPO substantially reduces computational costs (e.g., over 40\% savings in token consumption on the GPQA benchmark), while maintaining competitive accuracy across representative reasoning benchmarks.
\end{itemize}
By bridging ideas from test-time learning and reinforcement learning, we advance the broader goal of enabling LLMs to self-improve efficiently in dynamic environments.

%% file: sections/2-related-work.tex
\section{Related Work}
\label{sec:related-work}

\paragraph{Test-Time Adaptation for LLMs.}  
Traditional machine learning assumes a fixed training–testing separation, with frozen model parameters at inference. However, models often encounter distribution shifts at test time, leading to degraded performance. Early approaches to \emph{test-time adaptation} (TTA) proposed lightweight adjustments such as entropy minimization or updating normalization statistics on unlabeled test batches \citep{wang2020tent, niu2022efficient}. These methods demonstrated effectiveness on computer vision benchmarks but remain limited in expressiveness for large language models (LLMs), where domain shift can involve complex reasoning and linguistic variability.

\paragraph{Test-Time Learning and Training.}  
Recent work extends adaptation to full or partial parameter updates during inference. \citet{akyurek2024surprising} shows that test-time training (TTT) substantially improves the few-shot generalization of LLMs under distribution shift. \citet{hubotter2024efficiently} propose active fine-tuning strategies that selectively adapt LLMs using informative examples. Retrieval-based test-time training offers another perspective: \citet{hardt2023test} demonstrates that fine-tuning on nearest neighbors retrieved from a large corpus before answering each test instance (TTT-NN) significantly boosts language modeling performance. Similarly, \citet{hu2025test} proposes perplexity-based test-time adaptation for LLMs, framing inference as a continual self-supervised training process. Collectively, these works illustrate the broad potential of TTT in improving the adaptability of large models.

\paragraph{Test-Time Reinforcement Learning.}  
Reinforcement learning has also been employed for test-time training, with the incorporation of verification or feedback signals. \citet{li2025learning} introduces feedback-guided test-time training, where external verifiers provide weak supervision to refine predictions during inference. Moreover, \citet{zuo2025ttrl} presents Test-Time Reinforcement Learning (TTRL), which generates multiple rollouts, aggregates predictions through majority voting, and uses the consensus as a reward signal to update the policy. While effective, these methods suffer from inefficiency: they require many redundant rollouts to stabilize consensus estimates. More recently, \citet{wang2025reinforcement} shows that reinforcement learning with verifiable rewards (RLVR) can achieve strong reasoning gains with extremely few training examples, underscoring the potential of efficient RL in test-time settings. Our work builds on this line by explicitly addressing rollout inefficiency, introducing a statistically grounded early-stopping mechanism for adaptive allocation.

\paragraph{Policy Optimization in RL.}  
Policy optimization techniques have been widely used in LLM-based reinforcement learning. Proximal Policy Optimization (PPO) \citep{schulman2017proximal} is a standard algorithm for stabilizing policy updates under KL constraints and has been widely applied in reinforcement learning from human feedback. Group Relative Policy Optimization (GRPO) \citep{shao2024deepseekmath} extends this paradigm with group-normalized advantages, enabling more stable optimization for LLMs under limited feedback. 
OptPO leverages the generality of these surrogates, providing a plug-and-play rollout allocation mechanism that enhances efficiency without requiring changes to the underlying optimization algorithm.

In summary, prior work has established test-time training as a powerful framework for improving model adaptability during inference, spanning entropy minimization, verifier-guided learning, and reinforcement learning. However, existing test-time policy optimization methods remain computationally inefficient due to excessive rollout requirements. 
Our work fills this gap by introducing OptPO, a framework for optimal rollout allocation that substantially reduces compute costs while preserving performance, complementing and extending the existing test-time adaptation methods.

%% file: sections/4-methodology.tex
\section{Methodology}

\subsection{Preliminaries}
We consider test-time training for tasks with deterministic, verifiable targets (e.g., mathematics or science reasoning problems). Each test instance provides a prompt $x \in \mathcal{X}$, and the model $\pi_\theta(\cdot \mid x)$, parameterized by $\theta$, defines a stochastic policy over completions $y \in \mathcal{Y}$. Although ground-truth labels are unavailable at test time, we can estimate a self-consistent pseudo-label from multiple rollouts to guide adaptation.
The objective of test-time training is divided into two steps:
\textbf{(i) Pseudo-label estimation:} efficiently identify a consensus $a^\star$ via early-stopped majority voting; \textbf{(ii) Parameter update:} adapt $\pi_\theta$ using self-supervised feedback defined by $a^\star$, via either RL- or SFT-based updates.

\paragraph{Rollouts and Votes.}
Each completion $y$ is mapped to a canonical answer token $a = \mathcal{E}(y)$ through an \emph{answer extractor} $\mathcal{E}:\mathcal{Y}\to\mathcal{A}$, where $\mathcal{A}=\{1,\ldots,m\}$ denotes the finite set of possible answers. We model $m$ candidate hypotheses $\{H_1,\dots,H_m\}$, where $H_j$ represents the hypothesis that the correct answer equals $j$. Sequentially generating completions yields a sequence of \emph{votes}
\[
D_t = \mathcal{E}(Y_t), \qquad Y_t \sim \pi_\theta(\cdot \mid x),
\]
with $D_t \in \{1,\dots,m\}$ and cumulative vote history $D_{1:t} = \{D_1,\dots,D_t\}$. Each instance is allowed a maximum rollout budget $M$ and a minimum retained sample count $N \le M$. The goal is to form a reliable consensus pseudo-label $a^\star$ using as few rollouts as possible, then update $\theta$ at test time using a small set of $N$ retained rollouts.

\paragraph{Answer Probabilities.}
Conditioned on the true hypothesis $H_j$, each vote equals $j$ (the correct answer) with probability $p_0 \in (0,1)$, and one of the remaining $m-1$ candidates otherwise, with probabilities $\{p_r\}_{r=1}^{m-1}$ satisfying $p_0 + \sum_r p_r = 1$ and $p_0 > p_r$. Votes are conditionally independent given $H_j$. This categorical noise model captures the intuition that correct answers appear more often and provides a one-dimensional sufficient statistic for sequential testing based on vote-count gaps.

\subsection{Rollout Allocation with Probabilistic Optimality}
We treat each answer candidate as a hypothesis and perform a \emph{sequential probability ratio test} (SPRT) between the current leading and second-place classes as rollouts accumulate.

\paragraph{Likelihood and Posterior.}
Let $v_j(t)$ denote the number of votes for class $j$ up to time $t$, and let $L(t)$ and $S(t)$ denote the leading and runner-up classes. Under hypothesis $H_j$, the likelihood of the observed votes factorizes as
\[
P(D_{1:t}\mid H_j) \;=\; \prod_{k=1}^t P(D_k \mid H_j)
\;=\; p_0^{\,v_j(t)} \prod_{c\neq j} p_c^{\,v_c(t)}.
\]
With uniform priors $P(H_j)=1/m$, the posterior satisfies $P(H_j \mid D_{1:t}) \propto P(D_{1:t}\mid H_j)$.

\paragraph{Bayes Likelihood Ratio between Top Two Classes.}
We define the Bayes factor (likelihood ratio under uniform priors) as
\[
\Lambda_t \;=\; 
\frac{P(H_{L(t)} \mid D_{1:t})}{P(H_{S(t)} \mid D_{1:t})}
\;=\;
\frac{P(D_{1:t} \mid H_{L(t)})}{P(D_{1:t} \mid H_{S(t)})}.
\]
Expanding and simplifying the equation yields:
\[
\Lambda_t \;=\; 
\frac{p_0^{\,v_{L}(t)} \prod_{c\neq L} p_c^{\,v_c(t)}}
     {p_0^{\,v_{S}(t)} \prod_{c\neq S} p_c^{\,v_c(t)}}
\;=\;
\frac{p_0^{v_{L}(t)}}{p_0^{v_{S}(t)}} \cdot \frac{\big(p_{S\mid H_L}\big)^{v_{S}(t)}}{\big(p_{L\mid H_S}\big)^{v_{L}(t)}}.
\]
Crucially, the per-class error probabilities in the numerator and denominator are \emph{conditioned on different hypotheses}. Under $H_{L(t)}$, every incorrect class (including $S(t)$) shares the same probability mass. And symmetrically, under $H_{S(t)}$, the (now incorrect) leader $L(t)$ is one of the $m-1$ wrong classes, hence
\[
p_{S\mid H_L} \;=\; \frac{1-p_0}{m-1},
\qquad
p_{L\mid H_S} \;=\; \frac{1-p_0}{m-1}.
\]
Therefore $p_{S\mid H_L}=p_{L\mid H_S}=(1-p_0)/(m-1)$, and the extra factor simplifies to:
\begin{align*}
\frac{\big(p_{S\mid H_L}\big)^{v_{S}(t)}}{\big(p_{L\mid H_S}\big)^{v_{L}(t)}} 
&= \left(\frac{1-p_0}{m-1}\right)^{v_{S}(t)-v_{L}(t)} \\
&= \left(\frac{m-1}{1-p_0}\right)^{v_{L}(t)-v_{S}(t)}.
\end{align*}
Combining terms, the Bayes factor is derived as:
\[
\Lambda_t
= \left[\frac{p_0 (m-1)}{1-p_0}\right]^{\Delta_t}
\;\equiv\; \kappa^{\Delta_t},
\qquad
\kappa \;=\; \frac{p_0 (m-1)}{1-p_0},
\]
where $\Delta_t = v_{L}(t) - v_{S}(t)$ is the vote-count gap.

\paragraph{Stopping Criterion.}
Given user-specified type-I and type-II error budgets $\alpha,\beta\in(0,1)$, we define Wald thresholds as follows:
\[
A = \frac{1-\beta}{\alpha}, \qquad
B = \frac{\beta}{1-\alpha}, \qquad (A>1>B).
\]
At each step $t$, stop and select $L(t)$ if $\Lambda_t \ge A$, select $S(t)$ if $\Lambda_t \le B$, and otherwise continue (subject to the hard cap $M$). Since $\Lambda_t=\kappa^{\Delta_t}$ with $\Delta_t\in\mathbb{Z}_{\ge 0}$, we can equivalently threshold on the integer gap $\Delta_t$:
\[
\Delta_A \;=\; \left\lceil \frac{\log A}{\log \kappa} \right\rceil,
\qquad
\Delta_B \;=\; \left\lfloor \frac{\log B}{\log \kappa} \right\rfloor.
\]
If $\kappa>1$ (typical when $p_0 > \tfrac{1}{m}$), then $\Lambda_t$ increases with $\Delta_t$:
  stop for $L$ when $\Delta_t \ge \Delta_A$ and for $S$ when $\Delta_t \le \Delta_B$.
If $\kappa<1$ (pathological low-accuracy regime), the inequalities reverse; in practice we ensure $p_0>\tfrac{1}{m}$ so $\kappa>1$.

We denote the stopping time by
\[
\tau \;=\; \min\big\{\, N \le t \le M \,:\, \Delta_t \notin (\Delta_B,\Delta_A) \,\big\}.
\]
The final consensus pseudo-label is $a^\star = L(\tau)$ if $\Lambda_\tau \ge A$, or $a^\star = S(\tau)$ if $\Lambda_\tau \le B$; otherwise if no boundary is crossed by $M$, we set $a^\star=L(M)$.

This procedure adaptively determines how many rollouts each instance requires to reach a confident majority, forming the backbone of our efficient test-time training framework.

\paragraph{Rollout Allocation.}
Sampling proceeds until $t=\max\{N,\tau\}$, guaranteeing at least $N$ retained rollouts for learning while skipping redundant ones on easier instances. Saved compute can be reallocated to harder examples under a fixed global budget.

\subsection{Test-Time Policy Optimization}
Once a confident pseudo-label $a^\star$ is obtained, test-time adaptation updates the parameters $\theta$ based on the $N$ retained rollouts. The adaptation rule differentiates two paradigms depending on the update formulation.

\paragraph{RL-Based Optimization.}
For reinforcement-style updates, each retained rollout $(Y_i, a_i)$ receives a label-free reward measuring alignment with the consensus:
\[
R(Y_i; a^\star) \;=\; \mathbb{I}\!\left[a_i = a^\star\right],
\]
or more generally, $R(Y_i; a^\star)\in[0,1]$ for partial credit or verification heuristics. The test-time policy optimization objective for prompt $x$ is
\[
\max_{\theta} \;\; \mathbb{E}_{Y\sim \pi_\theta(\cdot \mid x)}[ R(Y; a^\star) ].
\]
With policy-gradient updates and baseline $b_i$, one on-policy step over $N$ samples is
\[
\theta \;\leftarrow\; \theta \;+\; \eta \frac{1}{N} \sum_{i=1}^{N} 
\nabla_\theta \log \pi_\theta(Y_i \mid x)\,\big(R(Y_i; a^\star) - b_i\big),
\]
optionally regularized to a reference policy $\pi_{\mathrm{ref}}$:
\begin{align}
\theta \;\leftarrow\; \arg\max_{\theta} \;
\frac{1}{N}\sum_{i=1}^{N}
\Big[
    \log \pi_\theta(Y_i \mid x)\,\hat{A}_i 
    \notag\\[-3pt]
    -\;
    \beta_{\mathrm{KL}}\,\mathrm{KL}\big(\pi_\theta(\cdot\mid x)\,\|\,\pi_{\mathrm{ref}}(\cdot\mid x)\big)
\Big],
\end{align}
where $\hat{A}_i = R(Y_i; a^\star) - b_i$. This objective is compatible with PPO \cite{schulman2017proximal}, GRPO \cite{shao2024deepseekmath}, or any general RL surrogates. The OptPO algorithm thus combines (i) BSPRT-based early stopping for efficient rollout allocation and (ii) on-policy policy optimization for rapid test-time adaptation.


\paragraph{SFT-Based Optimization.}
In the supervised variant, parameter updates are performed directly toward the pseudo-labels. For each batch of test inputs $\{x_i\}_{i=1}^B$, the model generates rollouts and determines a consensus label $G_i$ via majority voting. The standard test-time supervised fine-tuning (TTSFT) objective is:
\[
\mathcal{L}_{\mathrm{TTSFT}}(\theta)
= - \frac{1}{B} \sum_{i=1}^{B} \log \pi_\theta(G_i \mid x_i).
\]
To improve efficiency, our \emph{OptPO-SFT} improves the fixed-$N$ majority voting with the BSPRT-based early-stopping rule described earlier, producing adaptive pseudo-labels $G_i = L(\tau_i)$ where $\tau_i$ is determined by the likelihood ratio thresholds $(A,B)$. The same cross-entropy objective applies:
\[
\mathcal{L}_{\mathrm{OptPO-SFT}}(\theta)
= - \frac{1}{B} \sum_{i=1}^{B} \log \pi_\theta(G_i \mid x_i).
\]
Thus, OptPO-SFT shares the same update rule as TTSFT but allocates rollouts adaptively via the Bayesian stopping criterion, unifying efficiency with reliability.

\paragraph{Unified Perspective.}
Both RL-based and SFT-based variants of OptPO instantiate the same three-stage framework: They initiate \textit{rollouts}, compute \textit{BSPRT-based pseudo-label}, and finally perform \textit{parameter update}. They differ by the update rule: policy-gradient optimization versus cross-entropy fine-tuning, making the framework conceptually unified.

%% file: sections/5-experiments.tex
\section{Experiments}

\subsection{Experimental Setup}
\label{sec:experiment_setup}

\paragraph{Benchmarks.} 
We adopt challenging math and science reasoning benchmarks: 
(1) \textbf{MATH-500}~\citep{hendrycks2measuring}, with 500 high school competition problems; 
(2) \textbf{AIME 2024}~\citep{maa2024aime}, containing 30 pre-Olympiad level problems; 
(3) \textbf{AMC}, with intermediate-level problems from American math competitions. 
(4) \textbf{GPQA}~\citep{rein2024gpqa}, a Q\&A dataset of challenging, graduate-level science questions in biology, physics, and chemistry.

\begin{table*}[t!]
\centering
\caption{Performance comparisons of OptPO with existing test-time reinforcement learning methods.}
\label{tab:sota_results}
\setlength{\tabcolsep}{4pt} 
\footnotesize 
\begin{adjustbox}{width=\textwidth,center}
\begin{tabular}{l c c c c c c c c c c c c c c c c c c c c}
\toprule
{\textbf{Benchmark}} 
    & \multicolumn{5}{c}{\textbf{AIME 2024}} 
    & \multicolumn{5}{c}{\textbf{AMC}} 
    & \multicolumn{5}{c}{\textbf{MATH-500}} 
    & \multicolumn{5}{c}{\textbf{GPQA}}  \\
\cmidrule(lr){2-6} \cmidrule(lr){7-11} \cmidrule(lr){12-16} \cmidrule(lr){17-21}
    Metrics & \makecell{Mean\\@16} & \makecell{Pass\\@16} & \makecell{Pass\\@1} & Toks. & \makecell{Toks.\\saving} 
& \makecell{Mean\\@16} & \makecell{Pass\\@16} & \makecell{Pass\\@1} & Toks. & \makecell{Toks.\\saving}
& \makecell{Mean\\@16} & \makecell{Pass\\@16} & \makecell{Pass\\@1} & Toks. & \makecell{Toks.\\saving}
& \makecell{Mean\\@16} & \makecell{Pass\\@16} & \makecell{Pass\\@1} & Toks. & \makecell{Toks.\\saving}
\\
\midrule
\multicolumn{21}{l}{\textbf{Qwen2.5-Math-1.5B}} \\
\midrule
    TTRL-GRPO& 15.6& 43.3& 20.0& 142M& \multirow{2}{*}{17.25\%}& 47.7& 83.1& 54.2& 159M& \multirow{2}{*}{26.93\%}& 72.3& 90.6& 73.6& 218M& \multirow{2}{*}{32.57\%}& 26.5& 87.8& 30.5& 251M& \multirow{2}{*}{43.20\%}\\
    OptPO-GRPO& 14.6& 46.7& 20.0& 117M& & 48.5& 81.9& 54.2& 116M& & 72.3& 90.8& 73.8& 147M& & 26.8& 87.8& 34.5& 143M& \\
\cmidrule(lr){2-6} \cmidrule(lr){7-11} \cmidrule(lr){12-16} \cmidrule(lr){17-21}
    TTRL-PPO& 18.5& 46.7& 23.3& 143M& \multirow{2}{*}{15.20\%}& 48.1& 83.1& 55.4& 160M& \multirow{2}{*}{26.80\%}& 71.9& 91.2& 73.8& 231M& \multirow{2}{*}{35.01\%}& 25.9& 87.8& 29.9& 262M& \multirow{2}{*}{45.06\%}\\
    OptPO-PPO& 16.0& 46.7& 23.3& 121M& & 48.1& 83.1& 53.0& 117M& & 72.3& 91.0& 73.8& 150M& & 26.6& 89.8& 32.5& 144M& \\
\cmidrule(lr){2-6} \cmidrule(lr){7-11} \cmidrule(lr){12-16} \cmidrule(lr){17-21}
    TTRL-Rei++& 11.9& 43.3& 20.0 & 198M& \multirow{2}{*}{14.35\%}& 39.3& 81.9& 49.4& 242M& \multirow{2}{*}{27.42\%}& 55.9& 90.2& 56.4& 696M& \multirow{2}{*}{25.53\%}& 26.3& 93.9& 32.0& 431M& \multirow{2}{*}{43.52\%}\\
    OptPO-Rei++& 10.4& 46.7& 20.0& 170M& & 38.3& 81.9& 44.6& 176M& & 56.2& 89.8& 57.6& 518M& & 26.5& 90.9& 28.9& 243M& \\
\midrule
\multicolumn{21}{l}{\textbf{Llama-3.2-1B-Instruct}} \\ 
\midrule
    TTRL-GRPO& 3.5& 23.3& 6.7& 134M& \multirow{2}{*}{47.62\%}& 14.9& 51.8& 18.1& 128M& \multirow{2}{*}{78.36\%}& 30.3& 66.0& 30.8& 181M& \multirow{2}{*}{14.77\%}& 26.4& 82.2& 28.4& 134M& \multirow{2}{*}{44.74\%}\\
    OptPO-GRPO& 1.7& 20.0& 6.7& 70M& & 15.2& 55.4& 16.9& 28M& & 35.4& 64.4& 36.0& 155M& & 25.2& 81.2& 26.9& 74M& \\
\cmidrule(lr){2-6} \cmidrule(lr){7-11} \cmidrule(lr){12-16} \cmidrule(lr){17-21}
    TTRL-PPO& 3.8& 30.0& 6.7& 47M& \multirow{2}{*}{13.70\%}& 13.9& 55.4& 19.3& 75M& \multirow{2}{*}{41.76\%}& 31.2& 64.6& 31.6& 246M& \multirow{2}{*}{44.49\%}& 25.7& 83.2& 28.9& 159M& \multirow{2}{*}{46.19\%}\\
    OptPO-PPO& 3.8& 30.0& 6.7& 41M& & 17.1& 54.2& 22.9& 44M& & 34.1& 64.4& 34.8& 137M& & 26.7& 81.7& 31.0& 86M& \\
\cmidrule(lr){2-6} \cmidrule(lr){7-11} \cmidrule(lr){12-16} \cmidrule(lr){17-21}
    TTRL-Rei++& 1.5& 16.7& 6.7& 44M& \multirow{2}{*}{30.02\%}& 11.4& 54.2& 16.9& 32M& \multirow{2}{*}{18.96\%}& 33.4& 65.4& 32.2& 557M& \multirow{2}{*}{29.40\%}& 27.1& 85.8& 30.5& 478M& \multirow{2}{*}{41.96\%}\\
    OptPO-Rei++& 1.7& 23.3& 6.7& 31M& & 11.0& 51.8& 13.3& 26M& & 36.1& 66.6& 38.0& 393M& & 26.7& 85.8& 28.9& 278M& \\
\midrule
\multicolumn{21}{l}{\textbf{Qwen2.5-7B}} \\ 
\midrule
    TTRL-GRPO& 23.5& 40.0& 23.3& 87M& \multirow{2}{*}{38.56\%}& 52.1& 78.3& 54.2& 111M& \multirow{2}{*}{49.06\%}& 79.7& 91.0& 80.0& 172M& \multirow{2}{*}{45.02\%}& 35.1& 85.3& 39.1& 98M& \multirow{2}{*}{40.31\%}\\
    OptPO-GRPO& 23.5& 43.3& 23.3& 53M& &56.8& 77.1& 57.8& 56M& & 79.5& 91.6& 79.8& 94M& & 35.7& 84.8& 38.6& 59M& \\
\cmidrule(lr){2-6} \cmidrule(lr){7-11} \cmidrule(lr){12-16} \cmidrule(lr){17-21}
    TTRL-PPO& 24.4& 46.7& 23.3& 104M& \multirow{2}{*}{42.85\%}& 54.3& 79.5& 55.4& 101M& \multirow{2}{*}{41.66\%}& 80.3& 91.0& 80.4& 171M& \multirow{2}{*}{48.96\%}& 37.9& 83.8& 40.1& 109M& \multirow{2}{*}{39.83\%}\\
    OptPO-PPO& 24.0& 43.3& 26.7& 60M& & 53.2& 79.5& 55.4& 59M& & 78.1& 91.0& 78.4& 87M& & 36.8& 88.8& 39.1& 66M& \\
\cmidrule(lr){2-6} \cmidrule(lr){7-11} \cmidrule(lr){12-16} \cmidrule(lr){17-21}
    TTRL-Rei++& 20.8& 40.0& 23.3& 345M& \multirow{2}{*}{41.45\%}& 52.9& 80.7& 55.4& 430M& \multirow{2}{*}{43.50\%}& 78.9& 90.
    8& 80.0& 886M& \multirow{2}{*}{44.52\%}& 33.0& 85.3& 32.5& 139M& \multirow{2}{*}{45.47\%}\\
    OptPO+Rei++& 20.6& 40.0& 23.3& 202M& & 56.0& 79.5& 57.8& 243M& & 79.4& 91.0& 80.6& 491M& & 33.0 & 84.3& 34.0& 76M& \\
\bottomrule
\end{tabular}
\end{adjustbox}
\end{table*}

\begin{figure*}[t!]
    \centering
    \includegraphics[width=0.95\linewidth]{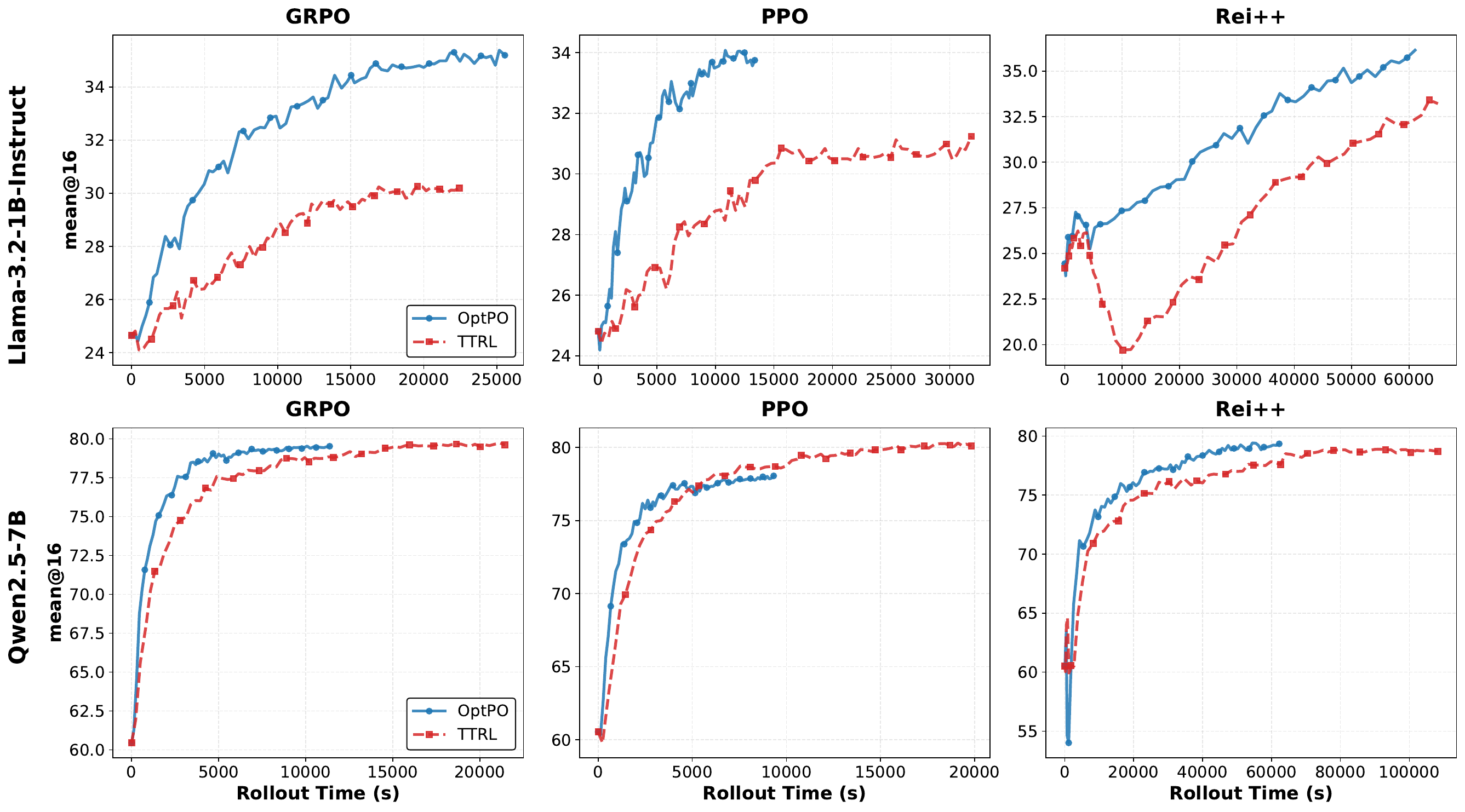}
    \caption{Performance (mean@16 accuracy vs rollout time) comparison of OptPO and TTRL on the MATH-500 benchmark, integrating different RL algorithms and using (\romannumeral1) Llama-3.2-1B-Instruct (top row) and (\romannumeral2) Qwen2.5-7B (bottom row) as backbones.}
    \label{fig:accuracy-time-figure}
\end{figure*}

\begin{table*}[t!]
\centering
\caption{Performance comparisons between OptPO-SFT and existing test-time supervised fine-tuning (TTSFT).}
\label{tab:sota_results_ttsft_optttsft}
\setlength{\tabcolsep}{4pt} 
\footnotesize 
\begin{adjustbox}{width=\textwidth,center}
\begin{tabular}{l c c c c c c c c c c c c c c c c c c c c}
\toprule
{\textbf{Benchmark}} 
    & \multicolumn{5}{c}{\textbf{AIME 2024}} 
    & \multicolumn{5}{c}{\textbf{AMC}} 
    & \multicolumn{5}{c}{\textbf{MATH-500}} 
    & \multicolumn{5}{c}{\textbf{GPQA}}  \\
\cmidrule(lr){2-6} \cmidrule(lr){7-11} \cmidrule(lr){12-16} \cmidrule(lr){17-21}
    Metrics & \makecell{mean\\@16} & \makecell{pass\\@16} & \makecell{pass\\@1} & toks. & \makecell{toks.\\saving} 
& \makecell{mean\\@16} & \makecell{pass\\@16} & \makecell{pass\\@1} & toks. & \makecell{toks.\\saving}
& \makecell{mean\\@16} & \makecell{pass\\@16} & \makecell{pass\\@1} & toks. & \makecell{toks.\\saving}
& \makecell{mean\\@16} & \makecell{pass\\@16} & \makecell{pass\\@1} & toks. & \makecell{toks.\\saving}
\\
\midrule
\multicolumn{21}{l}{\textbf{Qwen2.5-Math-1.5B}} \\
\midrule

    TTSFT& 13.1& 46.7& 20.0& 166M& \multirow{2}{*}{0.61\%}& 37.9& 80.7& 42.2& 138M& \multirow{2}{*}{12.08\%}& 59.7& 90.8& 63.8& 200M& \multirow{2}{*}{25.29\%}& 27.9& 81.8& 30.3& 113M& \multirow{2}{*}{51.17\%}\\
    OptPO-SFT& 13.5& 50.0& 20.0& 165M& & 39.7& 81.9& 43.4& 121M& & 59.0& 91.2& 60.2& 150M& & 27.6& 78.8& 31.3& 55M& \\

\midrule
\multicolumn{21}{l}{\textbf{Llama-3.1-8B-Instruct}} \\ 
\midrule

    TTSFT& 3.8& 26.7& 6.7& 314M& \multirow{2}{*}{15.33\%}& 20.1& 63.9& 28.9& 326M& \multirow{2}{*}{17.16\%}& 43.5& 81.2& 44.2& 653M& \multirow{2}{*}{19.84\%}& 28.8& 85.9& 30.3& 382M& \multirow{2}{*}{61.89\%}\\
    OptPO-SFT& 6.2& 36.7& 3.3& 266M& & 18.7& 60.2& 19.3& 270M& & 43.6& 82.8& 47.0& 523M& & 28.9& 82.8& 32.3& 146M& \\

\bottomrule
\end{tabular}
\end{adjustbox}
\end{table*}

\paragraph{Backbone Models.} 
We evaluate OptPO on widely used open-source LLM backbones, including Qwen2.5-Math-1.5B, Qwen2.5-7B, and Llama-3.2-1B-Instruct, to cover various model sizes and verify cross-architecture generalization. When verifying the effectiveness of TTSFT and OptPO-SFT, we utilized Qwen2.5-Math-1.5B and Llama-3.1-8B-Instruct as backbone models.

\paragraph{Baseline Methods.}
OptPO is a plug-and-play framework that can be plugged into multiple existing RL frameworks for improving test-time training efficiency. We plugged it into \textbf{PPO} \cite{schulman2017proximal}, \textbf{GRPO} \citep{shao2024deepseekmath}, and Reinforce++ (\textbf{Rei++}) \citep{hu2025reinforce} for the comprehensiveness of evaluation. We compare its performance with TTRL \citep{zuo2025ttrl}, the most prevalent test-time policy optimization method. We also evaluated the performance of OptPO-SFT in similar settings.



\paragraph{Implementation Details.}  Each rollout uses temperature 0.6 and nucleus sampling ($p=0.95$). We generate up to $M=64$ rollouts per instance with a minimum of $N=32$ or $N=16$ samples used for policy update. For RL-based updates (OptPO), we perform one policy-gradient step per instance using PPO/GRPO-style objectives with a KL penalty $\beta_{\mathrm{KL}}=0.001$. Actor Learning Rate is set to 1e-6, for RL algorithms involving a critic (like PPO), the critic learning rate is set to 1e-5. For SFT-based training (including TTSFT and OptPO-SFT), gradient descent is applied on the pseudo-labeled batch using AdamW with a learning rate of 2e-5.

The probability of answering a specific question $x$ correctly $p_0$ is estimated adaptively as the ratio of majority answer among the minimum retained rollouts, then times a degradation factor 0.6 to obtain a more realistic estimate of $p_0$, since instinctively pseudo-accuracy estimated by majority probability is higher than real accuracy. This calibration stabilizes BSPRT thresholds without labeled data. We set $(\alpha, \beta) = (0.05, 0.05)$, yielding a $95\%$ confidence level in the consensus decision. We require exceeding this confidence level 5 times across the rollouts for rigorous guarantees and robust early-stopping. All experiments are conducted on 8 $\times$ H20 GPUs.

\subsection{Main Results}

Table~\ref{tab:sota_results} reports the performance of our proposed methods across benchmarks and backbones. Overall, OptPO consistently outperforms its non-optimized counterpart TTRL, while substantially reducing rollout costs, demonstrating the effectiveness of our Bayesian early-stopping and adaptive pseudo-labeling mechanisms. Figure~\ref{fig:accuracy-time-figure} shows faster convergence and improvement of OptPO on the MATH-500 benchmark when using different backbone models. Our key findings are as follows: 

\paragraph{Substantial efficiency gains with comparable or superior accuracy.}
OptPO achieves comparable or higher mean@16 and pass@16 scores than baseline TTRL methods across all benchmarks, while consuming 30\% $\sim$ 50\% fewer tokens on average. For instance, on GPQA with Qwen2.5-Math-1.5B, OptPO+PPO achieves better accuracy in all metrics (mean@16, pass@16, and pass@1) than TTRL+PPO while reducing token usage from 262M to 144M, a 45\% compute saving. Similar trends are observed for other datasets and RL algorithms, confirming that our early-stopping scheme avoids redundant rollouts on easier samples and filters out harmful samples that occur due to no early-stopping. This ensures that efficiency is significantly improved without harming policy updates.

\paragraph{Cross-model and cross-algorithm generalization.}
Consistent improvements on both Qwen2.5 and Llama backbones indicate that our framework is architecture-agnostic. Even the smaller Llama-3.2-1B shows notable accuracy gains from OptPO-SFT and OptPO over baselines, highlighting that the proposed adaptation mechanism scales robustly across model sizes. Also, our OptPO framework integrates seamlessly with different types of RL algorithms (e.g., PPO, GRPO, Reinforce++) and consistently outperforms TTRL with higher accuracy and substantial efficiency gains, indicating its effectiveness across different RL algorithms.

\subsection{Evaluation on Test-time Fine-tuning}
Table~\ref{tab:sota_results_ttsft_optttsft} demonstrates our results for TTSFT and OptPO-SFT. Similar to RL-based variants, OptPO-SFT also achieves 15\% $\sim$ 60\% token consumption savings while maintaining similar or higher accuracy performance metrics than TTSFT.

\paragraph{SFT-based vs. RL-based variants.}
The RL-based OptPO variants generally yield stronger overall accuracy than the OptPO-SFT, particularly on math reasoning benchmarks. This suggests that policy-gradient adaptation better leverages stochastic feedback from rollouts compared to deterministic pseudo-label fine-tuning. Nevertheless, OptPO-SFT achieves more stable improvements over TTSFT across all settings, validating that our adaptive pseudo-labeling significantly enhances SFT-style test-time training.

\paragraph{Takeaways.}
Our results demonstrate that (1) adaptive early-stopping through BSPRT effectively improves efficiency without sacrificing accuracy; (2) OptPO offers the strongest balance between accuracy and compute; and (3) OptPO-SFT provides a lightweight alternative for rapid adaptation when RL-style updates are infeasible. Together, they establish a unified and efficient test-time training paradigm for verifiable reasoning tasks.

\input{sections/ablation_table}

\subsection{Ablation Studies}
\textbf{Ablation on the number of rollouts used for policy update $N$.} Table~\ref{tab:ablation_smaller_N} shows the performance of TTRL/OptPO when $N$ is 16. OptPO still demonstrates a consistent accuracy performance advantage over the TTRL baseline while saving 43.9\% of tokens than the TTRL baseline, which is even higher than 30.0\% saving when $N$ = 32. This shows the stability of the algorithm, which can still maintain great accuracy with amazing efficiency when we push the rollout budget to its limits.

\textbf{Ablations on type-I \& type-II error budgets $\alpha$ and $\beta$.} Table~\ref{tab:ablation_alpha_beta} shows the performance at OptPO under different type-I and type-II error budgets, when running using Qwen2.5-Math-1.5B on the MATH-500 benchmark. We find that OptPO's performance remains consistently strong and amazingly stable across all configurations. Notably, token savings will improve when we relax error tolerance to 0.07 and 0.1 without accuracy loss. This again confirms the stability and excellent performance of our OptPO framework.

%% file: sections/ablation_table.tex
\begin{table*}[th!]
\centering
\caption{Ablation studies on OptPO varying by $N$ for policy update using the Qwen2.5-Math-1.5B backbone.}
\label{tab:ablation_smaller_N}
\setlength{\tabcolsep}{4pt} 
\footnotesize 
\begin{adjustbox}{width=\textwidth,center}
\begin{tabular}{l c c c c c c c c c c c c c c c c c c c c}
\toprule
{\textbf{Benchmark}} 
    & \multicolumn{5}{c}{\textbf{AIME 2024}} 
    & \multicolumn{5}{c}{\textbf{AMC}} 
    & \multicolumn{5}{c}{\textbf{MATH-500}} 
    & \multicolumn{5}{c}{\textbf{GPQA}}  \\
\cmidrule(lr){2-6} \cmidrule(lr){7-11} \cmidrule(lr){12-16} \cmidrule(lr){17-21}
    \textbf{Metrics} & \makecell{Mean\\@16} & \makecell{Pass\\@16} & \makecell{Pass\\@1} & Toks. & \makecell{Toks.\\ Saving} 
& \makecell{Mean\\@16} & \makecell{Pass\\@16} & \makecell{Pass\\@1} & Toks. & \makecell{Toks.\\Saving}
& \makecell{Mean\\@16} & \makecell{Pass\\@16} & \makecell{Pass\\@1} & Toks. & \makecell{Toks.\\Saving}
& \makecell{Mean\\@16} & \makecell{Pass\\@16} & \makecell{Pass\\@1} & Toks. & \makecell{Toks.\\Saving}
\\
\midrule
    TTRL-GRPO ($N$=16) & 14.8 & 46.7 & 23.3 & 145M & \multirow{2}{*}{20.38\%} & 46.5 & 83.1 & 54.2 & 162M & \multirow{2}{*}{38.40\%} & 71.4 & 90.8 & 73.6 & 224M & \multirow{2}{*}{50.66\%} & 26.1 & 86.3 & 31.5 & 235M & \multirow{2}{*}{66.13\%}\\
    OptPO-GRPO ($N$=16)& 16.5 & 46.7 & 23.3 & 115M & & 46.2 & 83.1 & 55.4 & 100M & & 70.7 & 91.2 & 72.2 & 110M & & 27.1 & 88.3 & 31.5 & 79M & \\
\midrule
    TTRL-GRPO ($N$=32)& 15.6& 43.3& 20.0& 142M& \multirow{2}{*}{17.25\%}& 47.7& 83.1& 54.2& 159M& \multirow{2}{*}{26.93\%}& 72.3& 90.6& 73.6& 218M& \multirow{2}{*}{32.57\%}& 26.5& 87.8& 30.5& 251M& \multirow{2}{*}{43.20\%}\\
    OptPO-GRPO ($N$=32)& 14.6& 46.7& 20.0& 117M& & 48.5& 81.9& 54.2& 116M& & 72.3& 90.8& 73.8& 147M& & 26.8& 87.8& 34.5& 143M& \\
\bottomrule
\end{tabular}
\end{adjustbox}
\end{table*}

\begin{table}[h!]
\centering
\caption{Ablation studies on OptPO with different type-I ($\alpha$) and type-II ($\beta$) error budgets using Qwen2.5-Math-1.5B backbone on the MATH-500 benchmark.}
\label{tab:ablation_alpha_beta}

\footnotesize
\setlength{\tabcolsep}{3pt} 
\begin{adjustbox}{width=0.82\linewidth,center}
\begin{tabular}{l c c c c c}
\toprule
$\alpha$/$\beta$ & mean@16 & pass@16 & pass@1 & toks. & toks. saving \\
\midrule
0.01 & 72.0 & 90.8 & 74.8 & 146M & 32.90\% \\
0.03 & 72.1 & 90.4 & 74.2 & 147M & 32.69\% \\
0.05 & 72.3 & 90.8 & 73.8 & 147M & 32.57\% \\
0.07 & 71.9 & 91.0 & 73.8 & 141M & 35.26\% \\
0.1 & 72.2 & 90.6 & 73.8 & 141M & 35.25\% \\
\bottomrule
\end{tabular}
\end{adjustbox}
\end{table}

%% file: sections/6-conclusion.tex
\section{Conclusion}





This paper introduced OptPO, the first statistically grounded framework for efficient test-time policy optimization, which addresses the long-standing inefficiency issue of majority-vote–based rollouts in test-time reinforcement learning for LLMs. By formulating early-stopping consensus estimation as a Bayesian sequential probability ratio test, OptPO adaptively allocates rollouts based on posterior evidence rather than fixed budgets, ensuring reliable pseudo-labels while avoiding redundant sampling. The resulting framework is simple, plug-and-play, and compatible with standard policy optimization methods such as PPO, GRPO, and Reinforce++, as well as supervised test-time fine-tuning. Across diverse benchmark and backbone settings, OptPO consistently preserves or improves accuracy while achieving substantial compute savings. More broadly, this work underscores that efficient, label-free adaptation is both feasible and increasingly essential as LLMs are deployed in unpredictable real-world environments. We hope OptPO serves as a foundation for future research on principled, compute-efficient test-time learning.